\documentclass[a4paper,conference]{IEEEtran}
\IEEEoverridecommandlockouts
\ifCLASSINFOpdf
\else
\fi
\usepackage{graphicx}
\usepackage{booktabs}
\usepackage{color}
\usepackage{amssymb}
\usepackage{amsmath}
\usepackage{multirow}
\usepackage{booktabs}

\usepackage{amsfonts}
\usepackage{bm}
\usepackage{import}
\usepackage{cite}

\begin{document}
%
\title{Improving End-to-End Text Image Translation \\From the Auxiliary Text Translation Task}




\author{\IEEEauthorblockN{Cong Ma\textsuperscript{1,2},
Yaping Zhang\textsuperscript{1,2*}\thanks{*Corresponding author.},
Mei Tu\textsuperscript{4},
Xu Han\textsuperscript{1,2},
Linghui Wu\textsuperscript{1,2},Yang Zhao\textsuperscript{1,2}, and
Yu Zhou\textsuperscript{2,3}}
\IEEEauthorblockA{\textsuperscript{1}School of Artificial Intelligence, University of Chinese Academy of Sciences, Beijing 100049, P.R. China}
\IEEEauthorblockA{\textsuperscript{2}National Laboratory of Pattern Recognition (NLPR), Institute of Automation, Chinese Academy of Sciences,\\ No.95 Zhongguan East Road, Beijing 100190, P.R. China}
\IEEEauthorblockA{\textsuperscript{3}Fanyu AI Laboratory, Zhongke Fanyu Technology Co., Ltd, Beijing 100190, P.R. China\\
\IEEEauthorblockA{\textsuperscript{4}Samsung Research China - Beijing (SRC-B)}\\
Email: \{cong.ma, xu.han, linghui.wu, yaping.zhang, yang.zhao, yzhou\}@nlpr.ia.ac.cn, mei.tu@samsung.com}}
\maketitle

\begin{abstract}
End-to-end text image translation (TIT), which aims at translating the source language embedded in images to the target language, has attracted intensive attention in recent research. However, data sparsity limits the performance of end-to-end text image translation. Multi-task learning is a non-trivial way to alleviate this problem via exploring knowledge from complementary related tasks. 
In this paper, we propose a novel text translation enhanced text image translation, which trains the end-to-end model with text translation as an auxiliary task. 
By sharing model parameters and multi-task training, our model is able to take full advantage of easily-available large-scale text parallel corpus.
Extensive experimental results show our proposed method outperforms 
existing end-to-end methods, and the joint multi-task learning with both text translation and recognition tasks achieves better results, proving translation and recognition auxiliary tasks are complementary.~\footnote{Our codes are available at: \\https://github.com/EriCongMa/E2E\_TIT\_With\_MT.}
\end{abstract}


%
\IEEEpeerreviewmaketitle

\section{Introduction}
\begin{figure}[t]
	\centering
	\includegraphics[scale=0.45]{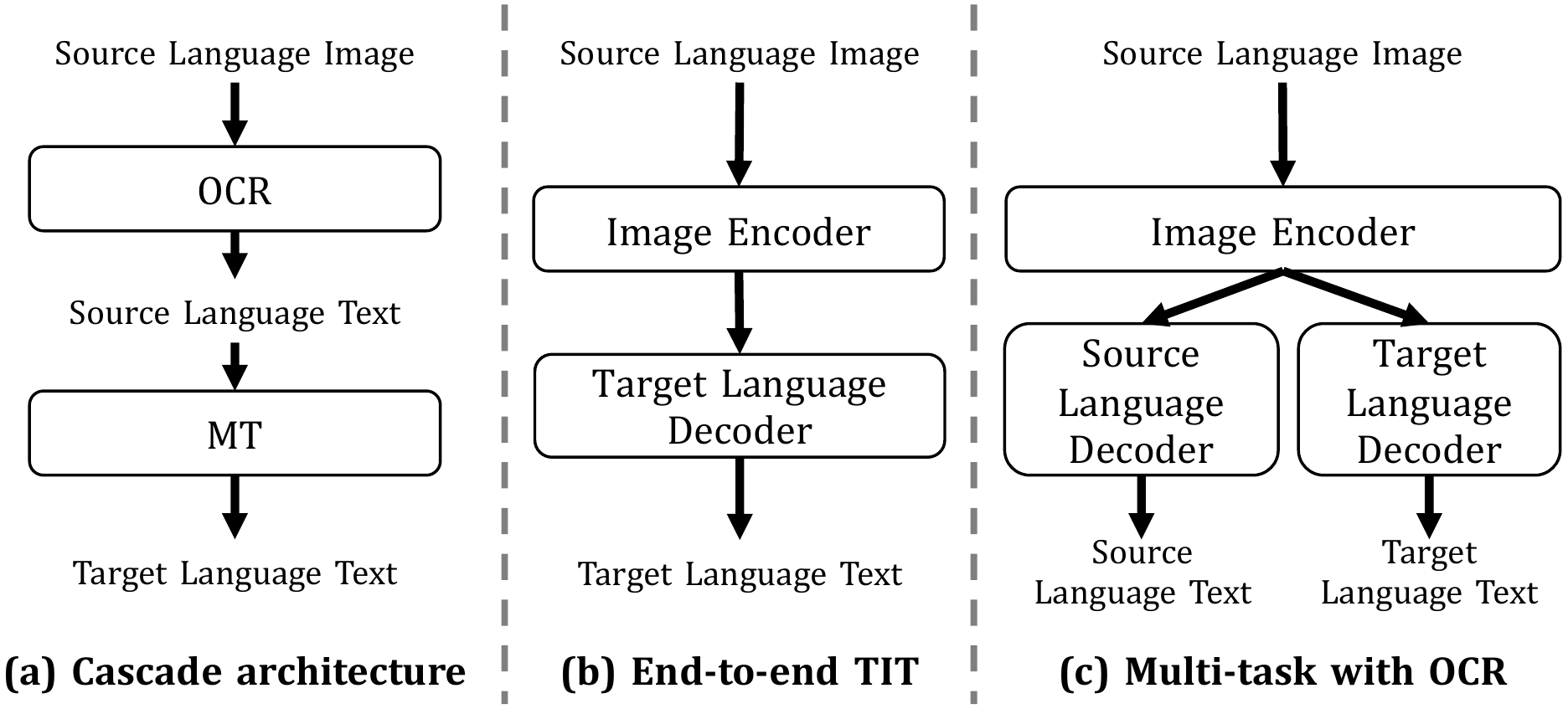}	
	\caption{Diagram of different architectures designed for text image translation.}
	\label{fig:overview_of_architectures}       
	\vspace{-0.25cm}
\end{figure}

Text image translation is widely used to translate images containing source language texts into the target language, such as photo translation, digital document translation, and scene text translation. 
Figure \ref{fig:overview_of_architectures} shows several architectures designed for TIT. Figure \ref{fig:overview_of_architectures} (a) depicts the cascade architecture, which utilizes optical character recognition (OCR) model and machine translation (MT) model together to translate source texts in images into target language ~\cite{manga_translation, Shekar2021OpticalCR, DBLP:conf/lt4dh/AfliW16, DBLP:journals/ijdar/ChenCN15, DBLP:conf/icdar/DuHSS11}. 
However, OCR models have recognition errors and MT models are vulnerable to noisy inputs. As a result, mistakes in recognition results are further amplified by the translation model, causing error propagation problems. Meanwhile, OCR and MT models are trained and deployed independently, which makes the overall progress redundant. The image containing the source language is first encoded and decoded by the OCR model, then it is encoded and decoded by the MT model, leading to high time and space complexity. In summary, the cascade architecture has the shortcomings of 1) error propagation, 2) parameter redundancy, and 3) decoding delay problems.


End-to-end architecture for TIT is shown as Figure~\ref{fig:overview_of_architectures} (b), which is designed to alleviate the shortcomings in cascade architecture by transforming text images into target language directly. However, end-to-end model training needs the dataset containing paired source language text images and corresponding translated target language sentences, which is difficult to collect and annotate, leading to data limitation problems. The existing methods utilized subtitle ~\cite{DBLP:conf/icpr/ChenYZYL20} or synthetic~\cite{DBLP:conf/icdar/SuLZ21} text line images to train and evaluate end-to-end models, but none of these datasets are released to the public, which limits the research and development of text image translation. Despite end-to-end text image translation data, both OCR and MT have large-scale available datasets, which are a valuable resource for text image translation. To train end-to-end model with external resource, existing methods explored to train end-to-end models with the help of auxiliary text image recognition task by incorporating external OCR dataset as shown in Figure~\ref{fig:overview_of_architectures} (c)~\cite{DBLP:conf/icpr/ChenYZYL20, DBLP:conf/icdar/SuLZ21}. 
However, multi-task training with OCR has two main limitations: 1) it only utilizes the OCR dataset, but ignores large-scale text parallel corpus; 2) OCR auxiliary task can only improve the image encoder optimization, but the decoder has not been fully trained.

\begin{figure*}[t]
	\centering
	\includegraphics[scale=0.79]{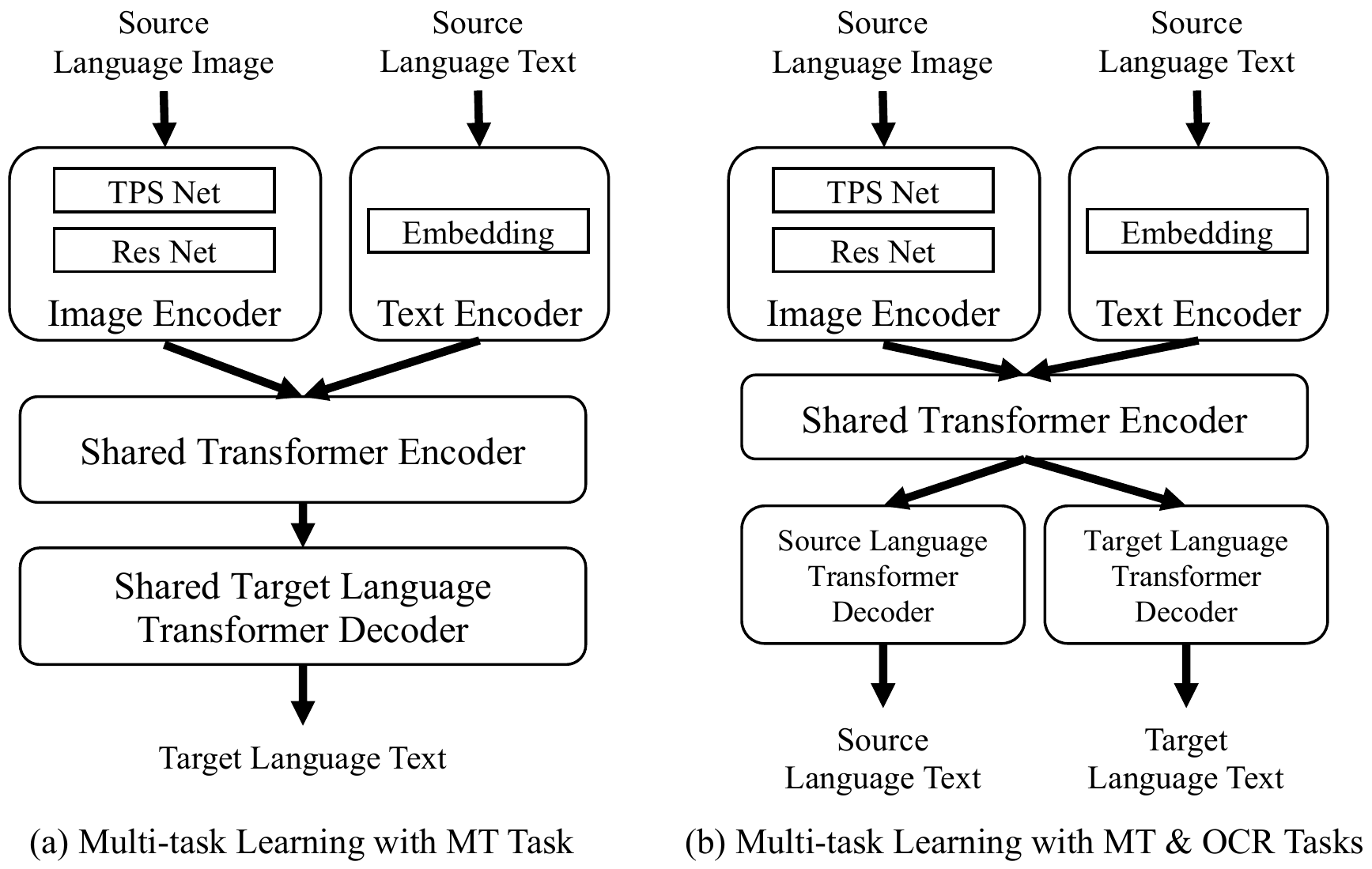}
	\caption{Architectures of multi-task learning for text image translation. Text encoder and source language transformer decoder are only utilized during training and abandoned when evaluation.}
	\label{fig:details_of_our_method}       
	\vspace{-0.25cm}
\end{figure*}


In order to address the shortcomings of multi-task training with OCR, we propose a novel text translation enhanced end-to-end text image translation model. 
As shown in Figure~\ref{fig:details_of_our_method} (a), multi-task learning is utilized to train TIT and MT tasks simultaneously. Shared parameters in the transformer encoder and decoder are fully trained with both image translation and text translation data. 
Specifically, images containing source language texts are encoded by the image encoder, and source language texts are encoded by the text embedding encoder separately. To learn the image features and text features into the same semantic feature space, a shared transformer encoder is utilized to encode both image and text features. Then, a shared transformer decoder for the target language generates translation results given semantic features from the transformer encoder. The shared transformer is optimized given both TIT and MT data, which has the potential to align image and text features into the shared feature space. Furthermore, By jointly training with MT and OCR tasks as shown in Figure~\ref{fig:details_of_our_method} (b), the end-to-end TIT model can take advantage of both external translation and recognition corpus. Since the recognition task generates source language texts, while the translation task generates target language texts, separated transformer decoders for source and target language are utilized to achieve OCR, MT, and TIT tasks under one unified architecture.

Extensive experiments on three translation directions show our method of multi-task training with MT auxiliary task outperforms multi-task training with OCR. Furthermore, by jointly training with MT and OCR tasks, our model achieves new state-of-the-art, proving that MT and OCR joint multi-task learning is complementary for text image translation. Meanwhile, our end-to-end model outperforms cascade models with fewer parameters and faster decoding speed, revealing the advantages of the end-to-end TIT model.





The contributions of our work are summarized as follows:

\begin{itemize}
\item We propose a multi-task learning architecture for text image translation with external text parallel corpus. Parameters in the semantic encoder and cross-lingual decoder are shared with the text translation model, which transfers the translation knowledge from the text translation task.
\item We further improve the end-to-end text image translation model by incorporating both MT and OCR auxiliary tasks, which improves the optimization of TIT, MT, and OCR tasks under one unified architecture.
\item Extensive experiments show our proposed method outperforms existing end-to-end methods by making better use of easily-available large-scale translation and recognition corpus. Meanwhile, our method outperforms cascade models with fewer parameters and faster decoding speed.

\end{itemize}

\section{Related Work}
\paragraph{OCR-MT Cascade System} OCR-MT cascade system first utilizes OCR model to recognize source language texts embedded in images~\cite{DBLP:journals/pami/ShiBY17, DBLP:conf/iccv/BaekKLPHYOL19, DBLP:conf/cvpr/ZhangNLXZS19, DBLP:conf/prcv/ZhangNLL19}. Then text translation models are incorporated to translate recognized source langauge texts into target language~\cite{DBLP:conf/nips/VaswaniSPUJGKP17, DBLP:conf/acl/GehringAGD17, DBLP:conf/nips/SutskeverVL14, DBLP:conf/ijcai/ZhaoZZZ20, DBLP:conf/coling/ZhaoXZZZZ20}.
In~\cite{manga_translation}, recognized sentences in manga images are considered as source language contexts for further document translation. Existing research also explores to integrate separated OCR and MT models for historical documents translation~\cite{DBLP:conf/lt4dh/AfliW16}, image document translation~\cite{DBLP:journals/ijdar/ChenCN15}, scene text translation~\cite{DBLP:conf/icassp/YangCZZW02} and photo translation on mobile devices~\cite{DBLP:conf/icdar/DuHSS11, DBLP:conf/icpr/WatanabeOKT98}. The cascade system can utilize large-scale datasets to train OCR and MT models independently. However, errors made by recognition models are propagated through MT models, which decreases the translation quality. Furthermore, the cascade system has two separated encoder-decoder architectures, leading to parameter redundancy and decoding delay problems.

\paragraph{End-to-End Text Image Translation} Mansimov et al. takes a preliminary step for end-to-end image-to-image translation without considering any text information~\cite{mansimov-etal-2020-towards}. Multi-task learning with OCR auxiliary task is utilized to alleviate the problem of data scarcity in end-to-end TIT task~\cite{DBLP:conf/icpr/ChenYZYL20}. To further utilize pre-trained models, a feature transformer is proposed to connect the pre-trained convolutional encoder in the OCR model and transformer decoder in the MT model for end-to-end TIT with OCR auxiliary multi-task learning~\cite{DBLP:conf/icdar/SuLZ21}. Although existing end-to-end methods explore improving text image translation performance with external OCR data, easily-available large-scale MT datasets are ignored. In this paper, we propose to train the end-to-end text image translation model with the MT auxiliary task, which incorporates text parallel corpus to improve the translation performance.



\section{Methodology}
\subsection{Problem Statement}
\label{label_problem_statement}
Text image translation is to translate source language text image into target language. Let $\bm{D}_\text{TIT}=\{\bm{I}, \bm{Y}\}$ be the text image translation corpus containing source language text image $\bm{I}$ and corresponding translation $\bm{Y}$. The end-to-end model optimizes the translation loss function of predicting target language sequence $\bm{Y}$ given image input $\bm{I}$:
\begin{equation}
\begin{aligned}
\mathcal{L}_{{E}}(\theta_{{E}})=-\sum_{(\bm{I},\bm{Y})\in\bm{D}_\text{TIT}}{\text{log}P({\bm{Y}}|\bm{I};\theta_{{E}})}
\end{aligned}
\end{equation}
where $\mathcal{L}_{{E}}$ and $\theta_{{E}}$ are the loss function and parameters of the end-to-end TIT models respectively.

\subsection{Backbone of Our Method}
The existing end-to-end models~\cite{DBLP:conf/icpr/ChenYZYL20, DBLP:conf/icdar/SuLZ21} are composed of an image encoder, a sequential modeling semantic encoder, and a decoder to map image pixels into the target language. Considering that text images have diverse text line directions, we strengthen the image encoder with a spatial transformation network, which is widely used in scene text recognition models.
As so, image encoder in our work consists of a Thin-Plate Spline (TPS) network~\cite{DBLP:conf/nips/JaderbergSZK15} and an CNN based feature extractor ResNet~\cite{DBLP:conf/cvpr/HeZRS16}.
Thin-plate spline, a variant of the Spatial Transformation Network (STN)~\cite{DBLP:conf/nips/JaderbergSZK15}, is utilized to normalize the input image by transforming the tilted texts in the image into the horizontal direction, which is essential for robust training and prediction. 
Specifically, TPS Net first detects the boundary of the texts in the image input. Second, text images of diverse directions are transformed into the horizontal position with a mapping function and a smooth spline interpolation, which releases the burden of extracting the tilted text image representation.
ResNet, which is composed of stacked convolution, pooling, and residual connection layers, is utilized to extract feature maps of the given image inputs. We implement the same architecture of ResNet which is used in \cite{DBLP:conf/iccv/ChengBXZPZ17}. This implementation has 29 trainable layers in total, and image features are extracted from the final convolutional layer of ResNet.
 
In summary, given the text image input $\bm{I}$, TPS Net first transforms it into the normalized image $\widetilde{\bm{I}}$, which has the horizontal text line in it. ResNet then transforms normalized image $\widetilde{\bm{I}}$ into feature maps with multiple convolutional layers and residual connections:
\begin{equation}
\begin{aligned}
F_\mathcal{I}&=\mathcal{I}(\bm{I};\theta_\mathcal{I})=\text{ResNet}(\text{TPS}(\bm{I}))
\end{aligned}
\end{equation}
where $F_\mathcal{I}$ denotes the image feature, and $\theta_\mathcal{I}$ represents parameters in image encoder.

\begin{table*}[t]
\centering
\setlength{\tabcolsep}{3.5mm}{
\caption{Statistics of text image translation (TIT), machine translation (MT), and optical character recognition (OCR) datasets.}
\label{stat_text_data}
\begin{tabular}{c|ccc|c|c|c|c}
\toprule[0.4mm]
\multirow{2}{*}{} & \multicolumn{3}{c|}{\textbf{Synthetic TIT Dataset}} & \textbf{Subtitle TIT Dataset} & \textbf{Street-view TIT Dataset} & \textbf{MT Dataset} & \textbf{OCR Dataset} \\
\specialrule{0em}{0.50pt}{0.50pt}
\cline{2-8}
\specialrule{0em}{0.50pt}{0.50pt}
& \#Train & \#Valid & \#Test & \#Test & \#Test & \#Train & \#Train \\
\specialrule{0em}{0.50pt}{0.50pt}
\hline
\specialrule{0em}{0.50pt}{0.50pt}
Zh$\Rightarrow$En & 1,000,000 & 2,000 & 2,502 & 1,040 & 1,198 & 5,984,287 & 1,000,000 \\
En$\Rightarrow$Zh & 1,000,000 & 2,000 & 2,502 & 1,040 & - & 5,984,287 & 1,000,000 \\
En$\Rightarrow$De 	  & 1,000,000 & 2,000 & 2,000 & - & - & 20,895,771 & 1,000,000 \\
\bottomrule[0.4mm]
\end{tabular}
}

\end{table*}

The text encoder is an embedding layer that maps text inputs into character embeddings. Specifically, given a text input $\bm{T}$, the text encoder first maps each character to an integer-based character index, and then utilizes a learnable linear projection to map the character index into text features:
\begin{equation}
	F_\mathcal{T}=\mathcal{T}(\bm{T};\theta_\mathcal{T})=\text{Embedding}(\bm{T})
\end{equation}
where $F_\mathcal{T}$ denotes the text feature, $\theta_\mathcal{T}$ denotes parameters in text encoder, and $\bm{T}$ represents source language sentence.

Shared transformer encoder $\text{TrE}(\cdot)$ is utilized to extract semantic features for text image translation and text translation tasks, while the shared target language transformer decoder $\text{TrD}(\cdot)$ is utilized to generate target translation.

\begin{equation}
\begin{aligned}
H_\mathcal{I}^{E}=\text{TrE}(F_\mathcal{I};\theta_\text{TrE})&;\ H_\mathcal{T}^E=\text{TrE}(F_\mathcal{T};\theta_\text{TrE})\\
H_\mathcal{I}^D=\text{TrD}(H_\mathcal{I}^E;\theta_\text{T-TrD})&;\ H_\mathcal{T}^D=\text{TrD}(H_\mathcal{T}^E;\theta_\text{T-TrD})\\
\end{aligned}
\end{equation}
where $H_\mathcal{I}^E$ and $H_\mathcal{T}^E$ represent transformer encoder output given image features and text features respectively. $H_\mathcal{I}^D$ and $H_\mathcal{T}^D$ represent transformer decoder output. $\theta_\text{TrE}$ represents parameters in shared transformer encoder, and $\theta_\text{T-TrD}$ represents parameters in shared target language transformer decoder. Please refer to~\cite{DBLP:conf/nips/VaswaniSPUJGKP17} for more details of the transformer.

Finally, the probability distribution of generating the target language is defined using a softmax layer:
\begin{equation}
\begin{aligned}
P(&\bm{Y}|\bm{I})= \text{Softmax}(W_oH_\mathcal{I}^D)\\
P(&\bm{Y}|\bm{T})= \text{Softmax}(W_oH_\mathcal{T}^D)\\
\end{aligned}
\end{equation}
where $W_o\in\mathbb{R}^{|\mathcal{V}_y|\times D}$ is a linear transformation mapping decoder features into target language space. $|\mathcal{V}_y|$ represents the size of target vocabulary, and $D$ is the hidden dimension of transformer output.


\subsection{Multi-task Learning with Text Translation Auxiliary Task}
Considering that text image translation is a cross-lingual generation task, we propose to train the end-to-end text image translation model with the MT auxiliary task, which is the core idea of our method. Given TIT data $\bm{D}_\text{TIT}=\{\bm{I}_i, \bm{Y}_i\}^{|\bm{D}_\text{TIT}|}_{i=1}$, the log-likelihood loss of end-to-end model can be formulated as follows:
\begin{equation}
\mathcal{L}_\text{TIT}(\theta_\mathcal{I},\theta_\text{TrE},\theta_\text{T-TrD})=-\sum_{i=1}^{|\bm{D}_\text{TIT}|}\text{log}P(\bm{Y}_i|\bm{I}_i)\\
\end{equation}

For text translation auxiliary task, denote $\bm{D}_\text{MT}=\{\bm{T}_i, \bm{Y}_i\}_{i=1}^{|\bm{D}_\text{MT}|}$ as the text parallel corpus containing bilingual sentences, the loss function of MT task is: 
\begin{equation}
\mathcal{L}_\text{MT}(\theta_\mathcal{T}, \theta_\text{TrE},\theta_\text{T-TrD})=-\sum_{i=1}^{|\bm{D}_\text{MT}|}\text{log}P(\bm{Y}_i|\bm{T}_i)
\end{equation}

\begin{figure*}[t]
	\centering
	\includegraphics[scale=0.54]{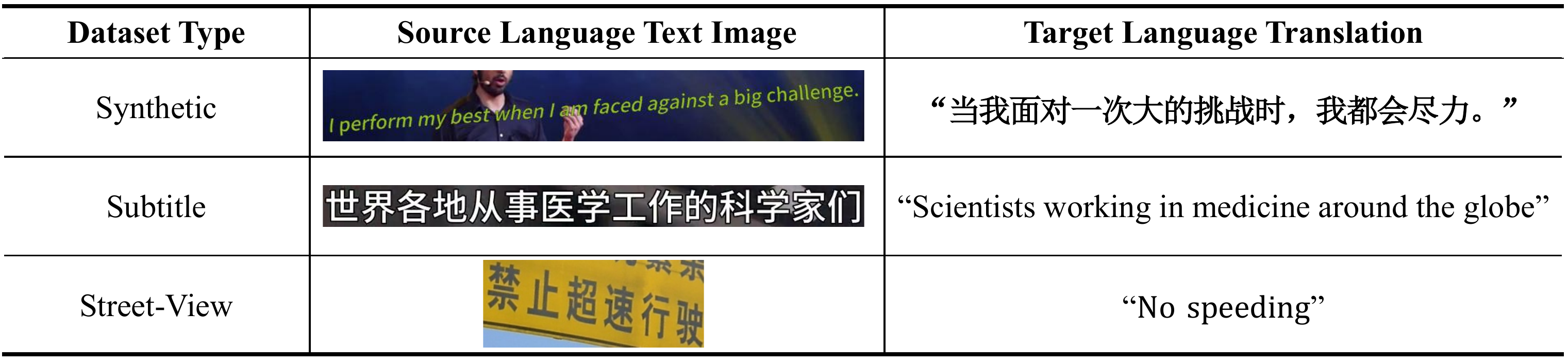}
	\caption{Examples of datasets utilized in our experiments.}
	\label{fig:examples_of_datasets}       
	\vspace{-0.5cm} 
\end{figure*}

By sharing the parameters $\theta_\text{TrE}$ and $\theta_\text{T-TrD}$ in semantic modeling transformer, multi-task learning with text machine translation optimizes the overall loss:
\begin{equation}
\begin{split}
\mathcal{L}&=\lambda_\text{TIT}\mathcal{L}_\text{TIT}+\lambda_\text{MT}\mathcal{L}_\text{MT}\\
\end{split}
\label{multitask_loss}
\end{equation}
where $\lambda_\text{TIT}$ and $\lambda_\text{MT}$ denote hyper-parameters to trade off two loss functions. During training, TIT and MT data are fed into models alternately to calculate corresponding losses and update the model parameters.


\subsection{Joint Multi-Task Learning with MT \& OCR Tasks}
Based on multi-task learning with MT auxiliary task, we further implement joint multi-task learning with both MT and OCR auxiliary tasks as shown in Figure~\ref{fig:details_of_our_method} (b). Decoders of OCR and MT tasks have different language settings, as a result, we split the shared transformer decoder into a source language transformer decoder for the OCR auxiliary task and a target language transformer decoder for the MT auxiliary task. Image encoder, text encoder, and shared transformer encoder are kept unchanged as multi-task learning with only MT auxiliary task. For OCR task, we have dataset as $\bm{D}_\text{OCR}=\{\bm{I}_i, \bm{T}_i\}^{|\bm{D}_\text{OCR}|}_{i=1}$ containing source language image $\bm{I}$  and source language text $\bm{T}$. The log-likelihood loss of the OCR task can be formulated as follows:
\begin{equation}
\mathcal{L}_\text{OCR}(\theta_\mathcal{I},\theta_\text{TrE},\theta_\text{S-TrD})=-\sum_{i=1}^{|\bm{D}_\text{OCR}|}\text{log}P(\bm{T}_i|\bm{I}_i)\\
\end{equation}

%

where $\theta_\mathcal{I}$, $\theta_\text{TrE}$, and $\theta_\text{S-TrD}$ denote parameters in image encoder, shared transformer encoder, and source language transformer decoder respectively. As a result, parameters in the shared transformer encoder  ($\theta_\text{TrE}$) are trained among all three tasks. OCR and TIT tasks share parameters in the image encoder ($\theta_\mathcal{I}$). MT and TIT tasks share parameters in target transformer decoder ($\theta_\text{T-TrD}$). The overall loss function of joint multi-task learning with both MT and OCR tasks is formulated as:
\begin{equation}
\begin{split}
\mathcal{L}&=\lambda_\text{TIT}\mathcal{L}_\text{TIT}+\lambda_\text{MT}\mathcal{L}_\text{MT} + \lambda_\text{OCR}\mathcal{L}_\text{OCR}\\
\end{split}
\end{equation}

where $\lambda_\text{TIT}$, $\lambda_\text{MT}$, and $\lambda_\text{OCR}$ are hyper-parameters to balance different tasks. Considering text image translation is the principle task, $\lambda_\text{TIT}$ is set to 1, and the weights of MT and OCR tasks are constraint as $\lambda_\text{MT}+\lambda_\text{OCR}=1$.

\begin{table*}[ht]
\centering
\setlength{\tabcolsep}{3.5mm}{
\caption{Performance of end-to-end models. All end-to-end models are trained with the same TIT training dataset. External OCR and MT corpus are also kept the same among different architecture settings.}
\label{tab:overall_experimental_results}
\begin{tabular}{c|c|l|ccc|cc|c}
\toprule[0.4mm]

\multirow{2}{*}{} & \multicolumn{2}{c|}{\multirow{2}{*}{Architecture}} & \multicolumn{3}{c|}{Synthetic}	 & \multicolumn{2}{c|}{Subtitle}  & {Street-view} \\
\specialrule{0em}{0.50pt}{0.50pt}
\cline{4-9}
\specialrule{0em}{0.50pt}{0.50pt}
& \multicolumn{2}{c|}{} & En$\Rightarrow$Zh & En$\Rightarrow$De & Zh$\Rightarrow$En & En$\Rightarrow$Zh & Zh$\Rightarrow$En & Zh$\Rightarrow$En \\
\specialrule{0em}{0.50pt}{0.50pt}\hline
\specialrule{0em}{0.50pt}{0.50pt}
1 & \multirow{2}{*}{CLTIR} & {End-to-End} & 18.02 & 15.55 & 10.74 & 16.47 & 9.04 & 0.43 \\
2 &  & End-to-End w/ OCR & 19.44 & 16.31 & 13.52 & 17.96 & 11.25 & 1.74 \\
\specialrule{0em}{0.50pt}{0.50pt}\hline
\specialrule{0em}{0.50pt}{0.50pt}
3 & \multirow{2}{*}{RTNet} & {End-to-End} & 18.91 & 15.82 & 12.54 & 17.63 & 10.63 & 1.07 \\
4 &  &End-to-End w/ OCR & 19.63 & 16.78 & 14.01 & 18.82 & 11.50 & 1.93 \\
\specialrule{0em}{0.50pt}{0.50pt}\hline
\specialrule{0em}{0.50pt}{0.50pt}
5 & \multirow{4}{*}{Our Work} & {End-to-End} & 19.25 & 16.27 & 13.16 & 17.73 & 10.79 & 1.69 \\
6 &  &End-to-End w/ OCR & 20.14  & 16.93 & 14.08 & 18.94 & 11.88 & 3.01 \\
7 &  & End-to-End w/ MT& {21.96} & {18.84} & {15.62} & {19.17} & \textbf{12.11} & {5.84} \\
8 &  & End-to-End w/ MT \& OCR & \textbf{22.13} & \textbf{19.06} & \textbf{15.69} & \textbf{19.46} & \textbf{12.11} & \textbf{5.86} \\

\bottomrule[0.4mm]
\end{tabular}}

\vspace{-0.2cm} 
\end{table*}

\section{Experiments}
\subsection{Dataset}
Since there is no publicly available dataset for end-to-end text image translation, we construct both synthetic and real-world datasets. OCR and MT datasets are also utilized in multi-task training. Examples are shown as in Figure~\ref{fig:examples_of_datasets}.

\paragraph{Text image translation dataset}
In order to implement the end-to-end text image machine translation model, we construct a synthetic text image dataset for training and annotate two real-world datasets for evaluation including subtitle and street-view test sets. Three language directions are considered in our work: English-to-Chinese (En$\Rightarrow$Zh), English-to-German (En$\Rightarrow$De), and Chinese-to-English (Zh$\Rightarrow$En). Synthetic text images are obtained with source language sentences in text parallel corpus and randomly sampled background images, which are collected from real-world video frames. Target language sentences are regarded as the ground truth of corresponding text image machine translation results. 
As shown in Table~\ref{stat_text_data}, we construct 1 million training data pairs of \{\textbf{source language image}, \textbf{target language text}\}, and 2,000 validation pairs for each language pair. The subtitle test set contains 1,040 samples, while the street-view test set has 1,198 samples. 

\paragraph{Text parallel dataset} 
Parallel text sentences in Workshop of Machine Translation\footnote{http://www.statmt.org/wmt18/} are utilized as the training corpus in text translation multi-task training. We obtain 5,984,287 En$\Leftrightarrow$Zh parallel sentences and 20,895,771 En$\Rightarrow$De parallel sentences. 

\paragraph{OCR dataset} To train text line recognition models in cascade system and multi-task learning with OCR auxiliary task, we synthesize 1 million English and Chinese text images respectively. Texts in images are randomly sampled from the monolingual source language corpus in MT datasets.

\subsection{Experimental Settings}
For model settings, the image encoder which includes TPS Net and Res Net is using the same configuration in ~\cite{DBLP:conf/iccv/BaekKLPHYOL19}. The text encoder represents a look-up embedding layer. The transformer is using the configuration of  transformer\_base in \cite{DBLP:conf/nips/VaswaniSPUJGKP17} which contains a 6-layer encoder and a 6-layer decoder with 512-dimensional hidden sizes. The maximum lengths for English, German and Chinese are set to 80, 80 and 40 respectively. Preprocessed image height is set to 32 and the channel is 3. To align the length of the image feature and text feature, preprocessed image width is resized to 320, 320 and 160. All models are trained with Adam optimizer~\cite{DBLP:journals/corr/KingmaB14} on a single NVIDIA V100 GPU. For evaluation metric, we report sacre-BLEU\footnote{https://github.com/mjpost/sacrebleu} on synthetic, subtitle, and street-view evaluation sets.

\subsection{Baseline Models}
\paragraph{CLTIR} In~\cite{DBLP:conf/icpr/ChenYZYL20}, the proposed end-to-end model is composed of CNN based network for image encoding, BiLSTM for sequential encoding, and attention-based gated linear unit (GLU) for target language decoding. Meanwhile, they trained end-to-end model with OCR auxiliary task. 
\paragraph{RTNet} In~\cite{DBLP:conf/icdar/SuLZ21}, the end-to-end model is also trained  with OCR auxiliary task. They utilized the transformer for sequential modeling and fine-tuned models by connecting the OCR encoder and MT decoder through a feature transformation module.

To provide a fair comparison, all the baseline models and our proposed method are trained with the same training dataset. OCR and MT datasets are also kept the same during multi-task learning.

%
%
%
	
\subsection{Results and Analysis}
\paragraph{Results of End-to-End Models} 
Table \ref{tab:overall_experimental_results} shows the results of end-to-end text image  translation on three evaluation datasets. For vanilla end-to-end training on the TIT training set (rows 1, 3 and 5), our end-to-end model (row 5) outperforms existing work due to the text image direction normalization module in our proposed image encoder. Incorporating OCR auxiliary task improves the performance (rows 2, 4 and 6), and our proposed multi-task with MT (row 7) outperforms all multi-task learning with OCR, indicating MT auxiliary task is vital to the text image translation task. Furthermore, jointly training with both OCR and MT tasks (row 8) achieves the best performance.

\begin{figure}[t]
	\centering

	\includegraphics[scale=0.56]{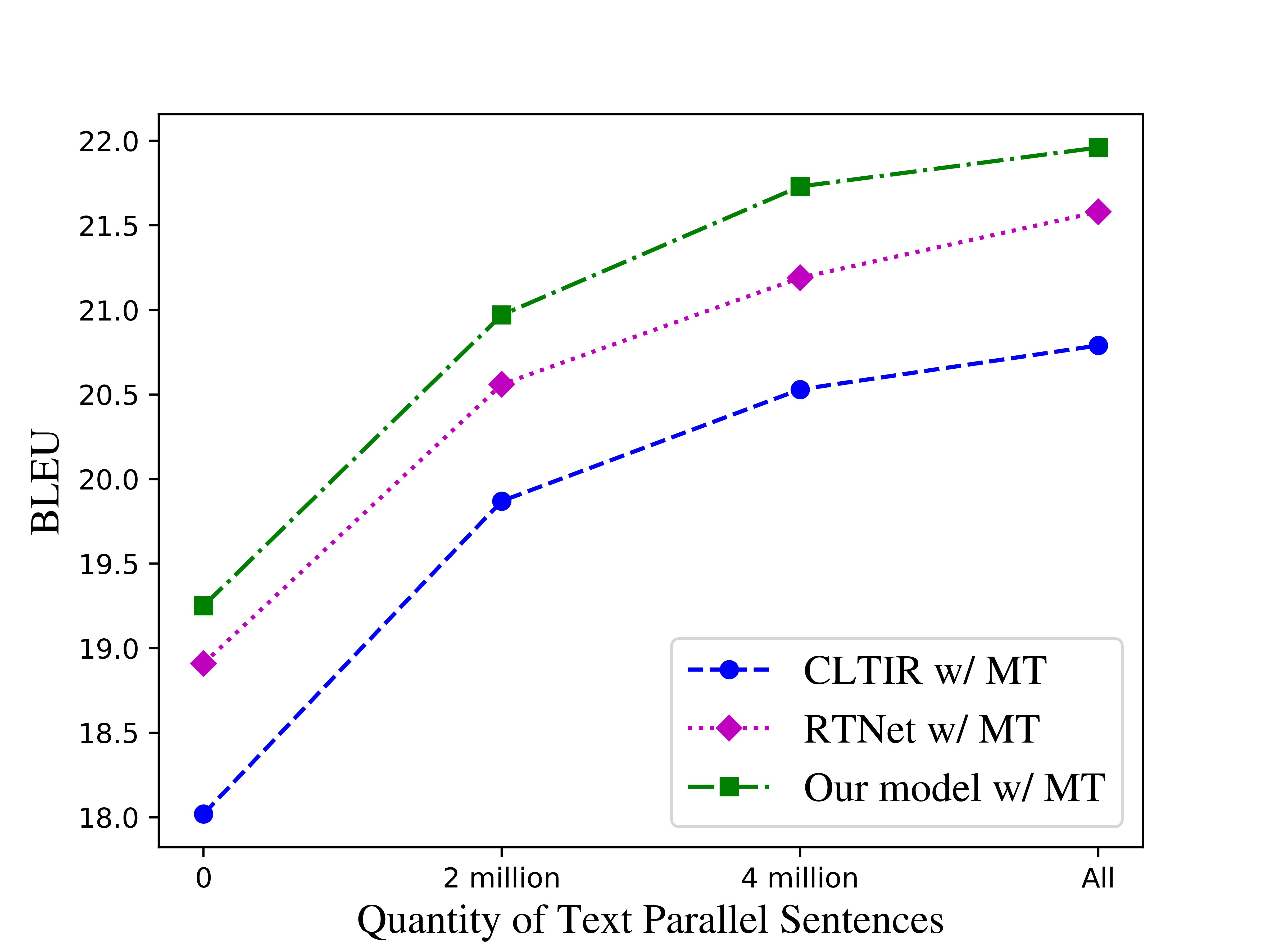}		
	\caption{Comparison of the quantity of text parallel corpus in English-to-Chinese translation direction.}
	\label{fig:analysis_of_quantity_parallel}       
\end{figure}

\paragraph{Effects of MT Auxiliary Task} 
To illustrate the effectiveness of multi-task learning with MT auxiliary task, we train both baseline and our models with MT auxiliary task. As shown in Figure~\ref{fig:analysis_of_quantity_parallel}, all models are improved after incorporating the MT task. The quantity of text parallel sentences influences the translation performance on a big scale, and the translation performance is continuously improved with the increased quantity of text parallel sentences, which proves the effectiveness of MT auxiliary task for text image machine translation.

\begin{table}[t]
\centering
\setlength{\tabcolsep}{6.5mm}{
\caption{Performance of OCR and MT models on the synthetic test set. Character Error Rate (CER) is utilized to evaluate the OCR model, while BLEU score is utilized to evaluate the MT model.}
\label{tab:cascaded_ocr}
\begin{tabular}{c|c|c|c}
\toprule[0.4mm]
\multicolumn{2}{c|}{CER of OCR Model} & \multicolumn{2}{c}{BLEU of MT Model} \\
\specialrule{0em}{0.50pt}{0.50pt}
\hline
\specialrule{0em}{0.50pt}{0.50pt}
\multirow{2}{*}{English} & \multirow{2}{*}{10.89\%} & En$\Rightarrow$Zh & 25.38 \\
& & En$\Rightarrow$De  & 20.97 \\
\specialrule{0em}{0.50pt}{0.50pt}
\hline
\specialrule{0em}{0.50pt}{0.50pt}
Chinese & 8.67\% & Zh$\Rightarrow$En  & 17.56 \\
\bottomrule[0.4mm]
\end{tabular}
}
	\vspace{-0.4cm} 
\end{table}

\begin{table}[h]
\centering
\setlength{\tabcolsep}{2.5mm}{
\caption{Comparison of cascade and end-to-end TIT models. The unit for parameters is million ($\times 10^6$), while the unit for decoding time is second per sentence.}
\label{tab:cascaded_e2e_enzh}
\begin{tabular}{l|l|l|l}
\toprule[0.4mm]
Architecture & BLEU & Parameters & Decoding Time \\
\specialrule{0em}{0.50pt}{0.50pt}
\hline
\specialrule{0em}{0.50pt}{0.50pt}
\multicolumn{4}{c}{English-to-Chinese Translation} \\
\specialrule{0em}{0.50pt}{0.50pt}
\hline
\specialrule{0em}{0.50pt}{0.50pt}
{Cascade} & 20.46 & 195M & 0.33 \\
{Our Method} & \textbf{22.13} ($\uparrow$1.67) & \textbf{122M} ($\downarrow$37.4\%) & \textbf{0.19} ($\downarrow$42.4\%) \\
\specialrule{0em}{0.50pt}{0.50pt}
\hline
\specialrule{0em}{0.50pt}{0.50pt}
\multicolumn{4}{c}{English-to-German Translation} \\
\specialrule{0em}{0.50pt}{0.50pt}
\hline
\specialrule{0em}{0.50pt}{0.50pt}
{Cascade} & 16.48 & 179M & 0.38 \\
{Our Method} & \textbf{19.06} ($\uparrow$2.58) & \textbf{114M} ($\downarrow$36.3\%) & \textbf{0.23} ($\downarrow$39.5\%) \\
\specialrule{0em}{0.50pt}{0.50pt}
\hline
\specialrule{0em}{0.50pt}{0.50pt}
\multicolumn{4}{c}{Chinese-to-English Translation} \\
\specialrule{0em}{0.50pt}{0.50pt}
\hline
\specialrule{0em}{0.50pt}{0.50pt}
{Cascade} & 15.12 & 225M & 0.21 \\
{Our Method} & \textbf{15.69} ($\uparrow$0.57) & \textbf{137M} ($\downarrow$39.1\%) & \textbf{0.11} ($\downarrow$47.6\%) \\
\bottomrule[0.4mm]
\end{tabular}
}

	\vspace{-0.2cm} 
\end{table}



\begin{figure}[t]
	\centering

	\includegraphics[scale=0.450]{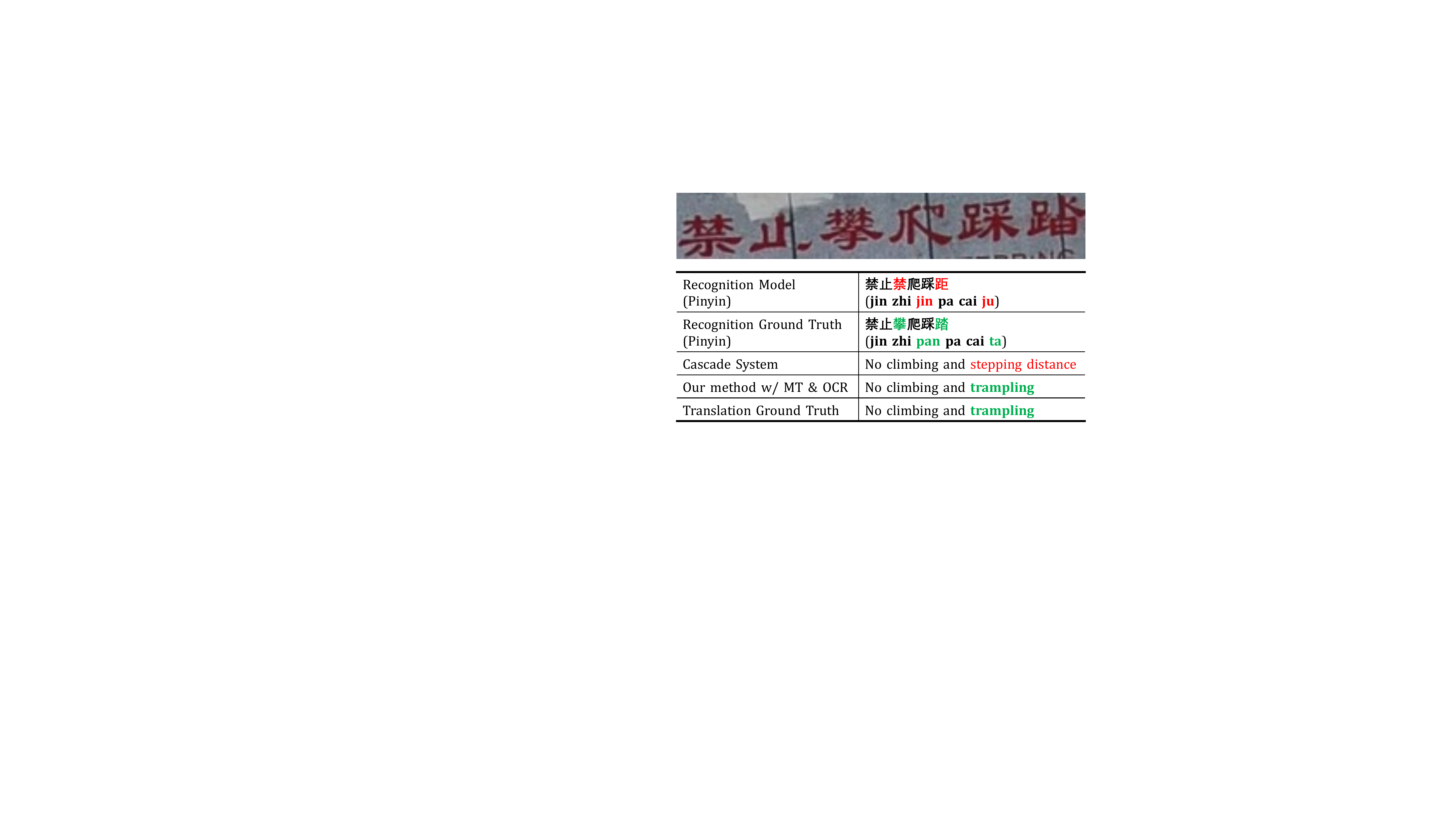}		
	\caption{Case Study of end-to-end and cascade text image translation.}
	\label{fig:case_study}       
\end{figure}

\begin{table}[t]
\centering
\setlength{\tabcolsep}{6.4mm}{
\caption{Ablation study of English-to-Chinese text image translation on synthetic valid set.}
\label{tab:ablation}
\begin{tabular}{l|c}
\toprule[0.4mm]
\multicolumn{1}{l|}{Architecture} & \multicolumn{1}{l}{BLEU} \\
\specialrule{0em}{0.50pt}{0.50pt}
\hline
\specialrule{0em}{0.50pt}{0.50pt}
\textbf{End-to-End w/ MT \& OCR} & \multicolumn{1}{l}{\textbf{24.23}} \\
\ {$-$ MT Auxiliary Task} & 23.17 ($\downarrow$ 1.06) \\
\ {$-$ OCR Auxiliary Task} & 22.25 ($\downarrow$ 0.92) \\
\ {$-$ TPS Net} & 21.44 ($\downarrow$ 0.81) \\
\ {$-$ Transformer} & 20.49 ($\downarrow$ 0.95) \\
\bottomrule[0.4mm]
\end{tabular}}
\end{table}

%

\begin{table}[!h]
\centering
\setlength{\tabcolsep}{2.75mm}{
\caption{Hyper-parameter evaluation of $\lambda_\text{MT}$ on English-to-Chinese synthetic valid set.}
\label{tab:hyper_parameter}
\begin{tabular}{c|cccccc}
\toprule[0.4mm]
$\lambda_\text{MT}$ & 0.0 & 0.2 & 0.4 & 0.6 & 0.8 & 1.0 \\
\specialrule{0em}{0.50pt}{0.50pt}
\hline
\specialrule{0em}{0.50pt}{0.50pt}
BLEU & 23.17 & 23.39 & 23.75 & \textbf{24.23} & 23.97 & 23.84 \\
\bottomrule[0.4mm]
\end{tabular}
}
	\vspace{-0.2cm} 
\end{table}



\paragraph{Comparison with Cascade System} 
Table ~\ref{tab:cascaded_ocr} shows the character error rate (CER) of OCR model and BLEU score of MT model in cascade system. OCR model is utilized the best model setting (TPS+ResNet+BiLSTM+Attn) in ~\cite{DBLP:conf/iccv/BaekKLPHYOL19}, and MT model is utilized transformer\_base setting in ~\cite{DBLP:conf/nips/VaswaniSPUJGKP17}. 
Table \ref{tab:cascaded_e2e_enzh} reveals that our proposed method outperforms the cascade system with fewer parameters and less decoding time in all language directions.
Through multi-task learning with both MT \& OCR auxiliary tasks, our proposed method improves translation performance by 1.61 BLEU scores on average, decreases about 37.6\% parameters, and spends around 43.2\% less time than the cascade system. Performance improvements in our method reveal that error propagation problems in the cascade system can be alleviated by end-to-end architecture with fewer parameters and less decoding time.

\paragraph{Ablation Study} The effectiveness of sub-modules in our method is analyzed in Table~\ref{tab:ablation}. After removing multi-task training with MT, performance drops significantly by 1.06 BLEU, while translation performance drops another 0.92 BLEU without OCR auxiliary multi-task learning. Without the TPS normalization module, the result decreases by 0.81 BLEU. 
The transformer is vital in modeling semantic information, which hurts the performance by  0.95 BLEU after replacing it with RNN based sequential module.

\paragraph{Hyper-parameter Analysis}
Hyper-parameters $\lambda_\text{TIT}$, $\lambda_\text{MT}$ and $\lambda_\text{OCR}$ in balancing TIT, MT and OCR losses are key parameters during model training. Considering TIT is the principle task in our experiments, we set $\lambda_\text{TIT}=1$ to make the model have enough information from text translation loss. Constraint of $\lambda_\text{MT}+\lambda_\text{OCR}=1$ is utilized during multi-task learning. We evaluate several hyper-parameter settings as shown in Table ~\ref{tab:hyper_parameter}. From this evaluation, the optimal value of $\lambda_\text{MT}$ is 0.6, and $\lambda_\text{OCR}=1-\lambda_\text{MT}=0.4$. With the increment of $\lambda_\text{MT}$, the performance drops due to the information reduction from OCR auxiliary task, proving MT and OCR auxiliary tasks are complementary in multi-task learning.

\paragraph{Case Study}
Figure~\ref{fig:case_study} shows one example that our proposed end-to-end method translates correctly, while the cascade system makes recognition errors and then obtains a bad translation. By jointly multi-task learning, our proposed method alleviates the shortcomings of error propagation in the cascade system.

\vspace{-0.25cm}

\section{Conclusion}
In this work, we present a novel machine translation enhanced end-to-end text image translation model. Extensive experimental results demonstrate training with text parallel corpus can significantly improve the TIT performance. Furthermore, jointly training with OCR and MT brings external improvements, indicating OCR and MT are complementary in multi-task learning for text image translation. 

Through our proposed method, multi-task learning can effectively address the data limitation problem in end-to-end TIT training by incorporating external MT and OCR data. Meanwhile, the end-to-end TIT model is proven effective and outperforms the cascade system with fewer parameters and less decoding time. 

\vspace{-0.25cm}

\section*{Acknowledgment}
This work has been supported by the National Natural Science Foundation of China (NSFC) grants 62106265.




%




\bibliographystyle{IEEEtran}
\bibliography{IEEEabrv,example}


\end{document}